\documentclass[journal,12pt,onecolumn,draftclsnofoot,]{IEEEtran}
\usepackage[left=2.07cm,right=2.07cm,top=2.4cm,bottom=2.7cm]{geometry}

\pdfpageattr{/Group << /S /Transparency /I true /CS /DeviceRGB >>}

\usepackage[utf8]{inputenc}
\usepackage[T1]{fontenc}
\usepackage{textcomp}
\usepackage{microtype}
\usepackage{calc}
\usepackage[normalem]{ulem}
\usepackage{verbatim}

\usepackage{lipsum}

\usepackage[range-phrase=--,per-mode=symbol-or-fraction,range-units=single,list-units=single,detect-all,forbid-literal-units=true]{siunitx}
\AtBeginDocument{
\DeclareSIUnit\bit{bit}
\DeclareSIUnit\byte{Byte}
\DeclareSIUnit\mbps{\mega\bit\per\second}
\DeclareSIUnit\kmh{\kilo\meter\per\hour}
\DeclareSIUnit\mw{\milli\watt}
\DeclareSIUnit\decibelm{dBm}
\DeclareSIUnit\decibeli{dBi}
\DeclareSIUnit\vehicle{veh}
}

\usepackage{silence}
\usepackage{csquotes}

\usepackage{amssymb,amsmath,amsthm}
\newtheorem{theorem}{Theorem}

\usepackage{amssymb}
\usepackage{amsfonts}
\usepackage{amsthm}
\usepackage{algpseudocode}

\usepackage{booktabs}
\usepackage{url}

\usepackage[inline]{enumitem}

\usepackage[caption=false,font=footnotesize]{subfig}
\usepackage[pdftex]{graphicx}
\DeclareGraphicsExtensions{.pdf,.png,.jpg}
\usepackage{tikz}
\usetikzlibrary{arrows}
\usetikzlibrary{calc}
\usetikzlibrary{chains}
\usetikzlibrary{scopes}

\usepackage[american]{babel}
\usepackage{hyphenat}

\usepackage{algorithm}
\usepackage[capitalize,noabbrev,nameinlink]{cleveref}

\RequirePackage{xstring}
\RequirePackage{xparse}
\RequirePackage[]{acro}
\makeatletter
\@ifpackagelater{acro}{2019/02/17}{
	\NewDocumentCommand\acrodef{mO{#1}mG{}}{\DeclareAcronym{#1}{short={#2}, long={#3}, foreign-plural={}, #4}}
}{
	\NewDocumentCommand\acrodef{mO{#1}mG{}}{\DeclareAcronym{#1}{short={#2}, long={#3}, #4}}
}
\makeatother

\acrodef{AI}{Artificial Intelligence}
\acrodef{BER}{Bit Error Rate}
\acrodef{CSI}{Channel State Information}
\acrodef{IoT}{Internet of Things}
\acrodef{LBT}{Listen-Before-Talk}
\acrodef{MCS}{Modulation and Coding Scheme}
\acrodef{NG-TCMS}{Next-Generation Train Control and Monitoring System}
\acrodef{NR}{New Radio}
\acrodef{QoS}{Quality of Service}
\acrodef{SCI}{Sidelink Control Information}
\acrodef{SCS}{Subcarrier Spacing}
\acrodef{SL}{Sidelink}
\acrodef{SNR}{Signal-to-Noise-Ratio}
\acrodef{SPS}{Semi-Persistent Scheduling}
\acrodef{RSRP}{Reference Signal Received Power}
\acrodef{TB}{Transport Block}
\acrodef{TSN}{Time-Sensitive Networking}
\acrodef{UE}{User Equipment}
\acrodef{V2I}{Vehicle-to-Infrastructure}
\acrodef{V2P}{Vehicle-to-Pedestrian}
\acrodef{V2V}{Vehicle-to-Vehicle}
\acrodef{V2X}{Vehicle-to-Everything}
\acrodef{WLCN}{WireLess Consist Network}
\acrodef{WLTB}{WireLess Train Backbone}
\acrodef{PRR}{Packet Reception Ratio}
\acrodef{PRB}{Physical Resource Block}
\acrodef{SINR}{Signal-to-Interference-plus-Noise-Ratio}
\acrodef{gNB}{g-NodeB}
\acrodef{LoS}{Line-of-Sight}

\acrodef{B.A.T.M.A.N.}{Better Approach To Mobile Ad-hoc Networking}
\acrodef{TTL}{Time to Live}
\acrodef{RB}{Resource Block}
\acrodef{OGM}{Originator Message}
\acrodef{MPR}{Multipoint Relaying}
\acrodef{PDR}{Packet delivery ratio}
\acrodef{D2D}{Device-to-Device}
\acrodef{OLSR}{Optimized Link State Routing Protocol}
\acrodef{AODV}{Ad hoc On-Demand Distance Vector Protocol}

\usepackage{todonotes}
\def\todoCtd#1{%
	TODO: #1%
	\ifx&#1&...\fi%
	\endgroup
	\relax
}

\NewDocumentCommand\IEEE{ s m >{\SplitArgument{4}{/}}d[] }{%
	\IfBooleanTF{#1}{}{IEEE\,}
	\nolinebreak[2]
	#2%
	\IfNoValueTF{#3}{%
	}{%
		\sommerIEEELettersSlashed#3%
	}%
}
\newcommand{\sommerIEEELettersSlashed}[5]{%
	\IfNoValueTF{#2}{%
	}{%
		\nolinebreak[3]
	}%
	#1%
	\IfNoValueTF{#2}{}{/#2}%
	\IfNoValueTF{#3}{}{/#3}%
	\IfNoValueTF{#4}{}{/#4}%
	\IfNoValueTF{#5}{}{/#5}%
}

\usepackage{pifont}

\usepackage[framemethod=tikz]{mdframed}
\newmdenv[
  linewidth=0.6pt,
  linecolor=black!20,
  backgroundcolor=black!2,
  roundcorner=3pt,
  innertopmargin=5pt, innerbottommargin=4pt,
  innerleftmargin=7pt, innerrightmargin=7pt,
  skipabove=6pt, skipbelow=6pt
]{specbox}

\begin{document}

\title{Understanding Learning Dynamics Through Structured Representations}

\author{%
\IEEEauthorblockN{%
    Saleh Nikooroo and Thomas Engel
}%

\small{
    \texttt{%
	saleh.nikooroo@uni.lu, thomas.engel@uni.lu
    }
}
\\
}

\maketitle

\begin{abstract}
While modern deep networks have demonstrated remarkable versatility, their training dynamics remain poorly understood—often driven more by empirical tweaks than architectural insight. This paper investigates how internal structural choices shape the behavior of learning systems. Building on prior efforts that introduced simple architectural constraints, we explore the broader implications of structure for convergence, generalization, and adaptation. 
Our approach centers on a family of enriched transformation layers that incorporate constrained pathways and adaptive corrections. We analyze how these structures influence gradient flow, spectral sensitivity, and fixed-point behavior—uncovering mechanisms that contribute to training stability and representational regularity. Theoretical analysis is paired with empirical studies on synthetic and structured tasks, demonstrating improved robustness, smoother optimization, and scalable depth behavior.
Rather than prescribing fixed templates, we emphasize principles of tractable design that can steer learning behavior in interpretable ways. Our findings support a growing view that architectural design is not merely a matter of performance tuning, but a critical axis for shaping learning dynamics in scalable and trustworthy neural systems.
\end{abstract}


\acresetall
\IEEEpeerreviewmaketitle

%


\section{Introduction}
\label{sec:introduction}

The design of neural network architectures has long balanced two competing goals: maximizing expressivity and maintaining trainability. While modern networks achieve state-of-the-art performance across a wide spectrum of tasks, the learning processes that unfold within them remain difficult to predict, interpret, or control. This opacity has motivated a growing interest in understanding how architectural decisions influence the underlying dynamics of training—particularly in settings where depth, data scarcity, or instability pose practical limitations.

Recent efforts have explored how different forms of structure—ranging from symmetry constraints to spectral regularization—can shape training behavior. Yet, much of this work remains either heuristic or narrowly tied to specific tasks or domains. In this paper, we pursue a broader goal: to investigate how architectural structure, even when modest or non-physical in origin, can systematically influence core properties of learning, such as gradient propagation, convergence rates, and functional smoothness.

Building on recent insights that link internal pathways to stability and signal coherence, we explore a class of structured transformations that decouple primary signal shaping from adaptive correction. While prior work has demonstrated empirical benefits in training stability and robustness, the present study takes a second look—with an emphasis on theoretical framing, diagnostic interpretation, and a more refined understanding of the mechanisms at play.

Our contributions are threefold:
\begin{itemize}
    \item We formalize the impact of structured transformation pathways on gradient dynamics and spectral behavior, providing a clean lens to understand their regularizing influence.
    \item We propose analytical tools and empirical probes that reveal how structural constraints alter internal representations, affect convergence geometry, and suppress pathological learning regimes.
    \item We present experiments on synthetic and graph-based settings, illustrating that well-chosen architectural structure can improve both robustness and generalization, even without explicit inductive biases or domain knowledge.
\end{itemize}

Through this work, we aim to clarify how architectural constraints can serve not just as stabilizers or performance boosters, but as active agents in shaping the flow and evolution of learning itself. Rather than viewing structure as a limitation, we treat it as a means to uncover new pathways for understanding and improving the deep learning process.

%

\section{Architectural Influence on Learning Dynamics}
\label{sec:architectural_dynamics}

The success or failure of training a neural network often hinges on the internal geometry induced by its architecture. While optimization algorithms operate externally, the curvature, conditioning, and propagation of gradients are deeply shaped by how transformations are arranged and parameterized. This section investigates how specific architectural choices—particularly structured transformations and correction paths—affect key aspects of learning dynamics.

\subsection{Preliminaries and Setup}

We consider a feedforward model composed of $L$ layers, each implementing a transformation of the form:

\begin{equation}\label{eq:FF_model}
x^{(l)} = \underbrace{\mathcal{S}^{(l)} W^{(l)} x^{(l-1)}}_{\text{Structure-Aware Path}} + \underbrace{\mathcal{C}^{(l)}(x^{(l-1)})}_{\text{Adaptive Correction Path}},
\end{equation}

\noindent where $\mathcal{S}^{(l)}$ is a shaping operator that constrains or modulates $W^{(l)}$, and $\mathcal{C}^{(l)}$ is a learnable correction function. This decomposition generalizes standard affine layers by allowing explicit architectural constraints to act before optimization begins. We refer to this structured–corrective architecture as a {Physics-Guided Neural Network (PGNN)}.

\subsection{Gradient Flow and Spectral Conditioning}

Let $\mathcal{L}$ be the loss function, and consider the gradient $\nabla_{x^{(l)}} \mathcal{L}$. In deep networks, this quantity often suffers from amplification or attenuation due to repeated matrix products. The introduction of $S^{(l)}$ effectively regularizes the spectral properties of each transformation:
\begin{equation}
    \sigma_{\text{max}}(S^{(l)} W^{(l)}) < \sigma_{\text{max}}(W^{(l)}),
\end{equation}
for appropriately chosen $S^{(l)}$. As a result, gradient norms remain better conditioned across layers, reducing the likelihood of vanishing or exploding behavior.

\paragraph*{Directional bias}
Structured paths bias gradient energy toward dominant modes of $\mathcal{S}^{(l)}$.
We evidence this via Jacobian-spectrum and gradient-norm diagnostics (Sec.~\ref{sec:fmnist})
and model-derived CKA/Subspace-Overlap (Sec.~\ref{sec:mechanistic}).

\subsection{Corrective Paths as Dynamic Modulators}

The correction term $\mathcal{C}^{(l)}$ plays a dual role: it preserves model expressivity and absorbs residual error that structured components cannot capture. Unlike architectural mechanisms such as batch normalization or residual connections, $\mathcal{C}^{(l)}$ does not merely pass identity or rescale activations—it actively learns to modulate the shaped transformation.

This dynamic modulation can be interpreted as a kind of learned response function that adapts to compensate for rigidity in $\mathcal{S}^{(l)} W^{(l)}$. As such, the interaction between structured and corrective paths forms a two-way system: one enforces stable propagation, the other flexibly recovers expressivity.

\subsection{Implications for Depth and Robustness}

One of the primary benefits of this formulation is improved depth scalability. By ensuring that each transformation respects spectral constraints and directional smoothness, the model can be stacked deeper without requiring skip connections or aggressive normalization. Moreover, the coupling with $\mathcal{C}^{(l)}$ provides a built-in mechanism for local adaptation, mitigating the risks of structural bias or underfitting.

These effects are further explored in our empirical section, where we study training dynamics under perturbations, depth variation, and projection ablations.


%

\section{Related Work}
\label{sec:related}

A growing body of research has explored how architectural structure influences the dynamics and generalization of neural networks. Our work builds on this line of inquiry by proposing a compositional framework that combines structured transformations with adaptive correction paths—offering new tools for analyzing and guiding the learning process. In our recent work \cite{nikooroo2025_structure_transform}, we introduced a dual-stage neural architecture in which each layer is decomposed into a structured transformation followed by an adaptive corrective term. This formulation, grounded in spectral and geometric constraints, enables provable improvements in gradient conditioning, training stability, and interpretability—particularly in sparse or low-rank regimes. By embedding architectural priors directly into the forward computation, the structured-corrective design enhances robustness without compromising expressivity. The present paper builds on this paradigm by developing a compositional framework that generalizes corrective paths across space, depth, and semantic roles—laying the groundwork for more modular and adaptive neural systems.

A. Saxe et al. introduce the Gated Deep Linear Network framework to explain how information pathways impact dynamics and generalization in multitask and transfer learning settings \cite{saxe2022neural}, offering a high-level abstraction for shared representation learning.

B. Baker provides a detailed analysis of how architectural constraints and activation linearity affect low-rank gradient dynamics and bottleneck collapse in deep networks \cite{baker2024lowrank}, with implications for stable and efficient training.
A. Chen et al. propose Structured Neural Networks (StrNN) as a method for injecting architectural inductive bias into generative and causal inference models \cite{chen2023structured}, demonstrating data efficiency gains via built-in structural constraints.

C. Lyle et al. explore how plasticity loss in neural networks can be mitigated through architectural design and normalization strategies \cite{lyle2024plasticity}, arguing that structure serves not just stability but long-term adaptability.

Z. Gao introduces a self-regularized graph neural network (SR-GNN) that improves stability by constraining frequency responses in the spectral domain \cite{gao2022spectral}, highlighting how structural filtering supports robust optimization in non-Euclidean domains.

E. Boursier et al. provide a fine-grained analysis of gradient flow in shallow ReLU networks with orthogonal inputs \cite{boursier2022gradient}, establishing convergence guarantees under minimal initialization and clarifying optimization geometry.

L. Wu and collaborators show that stochastic gradient descent (SGD) implicitly regularizes networks through dynamical stability, outperforming vanilla gradient descent in generalization, especially in low-depth settings \cite{wu2023implicit}.
F. Sherry et al. propose ResNet-inspired architectures that encode non-expansive operators via spectral norm constraints, leading to stable and adversarially robust networks \cite{sherry2023stable}.

W. Chen et al. analyze how deep connectivity patterns influence convergence under gradient descent, using graph-theoretic tools to identify and prune ineffective pathways \cite{chen2022connectivity}.

A. Damian et al. develop the concept of self-stabilization, showing that gradient descent at the edge of stability follows a form of projected optimization constrained by sharpness \cite{damian2022self}, with theoretical and empirical support across tasks.
M. Xu et al. study square-loss deep classifiers and reveal how normalization, low-rank evolution, and neural collapse emerge as beneficial dynamics, particularly in sparse or modular architectures \cite{xu2023dynamics}.

Further recent work continues to highlight how structural and dynamical properties shape deep learning outcomes.
B. Dherin et al. analyze the gradient flow of networks trained with square loss and orthogonal inputs, showing that common training heuristics influence geometric complexity and double-descent behavior \cite{dherin2022geometry}.
C. Lyle et al. explore how changes in the curvature of the loss landscape affect plasticity in deep networks, linking architectural and optimization choices to preserved adaptability during training \cite{lyle2023plasticity}.

L. Noci et al. study signal propagation in Transformers, demonstrating how architectural constraints can mitigate rank collapse and balance gradient flow \cite{noci2022transformers}.
F. Chen et al. propose the concept of stochastic collapse, where gradient noise biases SGD toward simpler subnetworks—enhancing generalization by aligning optimization with model sparsity \cite{chen2023stochastic}.

A. Gravina et al. introduce Anti-Symmetric Deep Graph Networks (A-DGNs), a stable architecture that preserves long-range dependencies and avoids vanishing/exploding gradients in deep graph models \cite{gravina2022antisymmetric}.

S. Mittal et al. assess modular neural architectures, finding that while modularity improves generalization and interpretability, it may require additional inductive biases or design principles to achieve full specialization \cite{mittal2022modular}.

Recent studies have also explored how structural priors interact with self-supervised learning, symbolic reasoning, and Transformer dynamics—further emphasizing the role of architectural constraints in shaping learning outcomes.

V. Cabannes et al. analyze how data augmentations, inductive bias, and optimization interact in self-supervised learning \cite{cabannes2023ssl}, proposing a theoretical framework that links training procedures to generalization bounds in domain-shifted tasks.
R. Riedi et al. study singular value perturbations in deep networks, showing how architectural choices influence optimization surfaces and lead to better-conditioned landscapes \cite{riedi2022singular}, especially in skip-connected architectures like ResNets.

D. Campbell et al. demonstrate that relational constraints in network architecture can reproduce human-like biases toward geometric regularity, offering an alternative to symbolic reasoning for structure discovery tasks \cite{campbell2023relational}.
Y. Li et al. provide a mechanistic view of how Transformers learn semantic topic structures, revealing that embedding and self-attention layers encode co-occurrence patterns in complementary ways \cite{li2023topic}.
E. Nichani et al. show that Transformers can learn latent causal graphs through gradient descent by encoding structure directly into the attention matrix, leading to robust generalization across reasoning tasks \cite{nichani2024causal}.

Further contributions have continued to reveal how architecture and implicit dynamics shape learning efficiency, generalization, and robustness across various neural settings.

N. Razin et al. study implicit regularization in hierarchical tensor factorizations, showing that architectural depth in convolutional networks induces regularization toward locality and non-trivial inductive structure \cite{razin2022implicit}.
D. Teney et al. challenge the assumption that random networks behave like random functions, demonstrating that structural components like ReLU and residual connections lead to consistent biases and predictive performance \cite{teney2024redshift}.
M. S. Nascon et al. investigate the role of step size in linear diagonal networks, revealing that large step sizes induce sparsity and directional stability—serving as a form of implicit architectural bias \cite{nascon2022stepbias}.

F. Di Giovanni et al. examine the over-squashing problem in message-passing networks, showing that width, depth, and graph topology strongly influence the ability of neural networks to propagate signals effectively \cite{giovanni2023oversquashing}.
S. Tang et al. explore the connection between architectural design and adversarial robustness, highlighting how sparsity and attention structures in Vision Transformers improve generalization over convolutional baselines \cite{tang2022robust}.
K. Kawaguchi et al. analyze the information bottleneck in deep networks, providing a theoretical link between compression at hidden layers and reduced generalization error, especially under excess representation \cite{kawaguchi2023bottleneck}.

Together, these works reinforce the broader perspective that stability, modularity, and inductive constraints are not peripheral design considerations—they fundamentally shape how neural systems learn, generalize, and adapt. Our structured-corrective architecture contributes to this conversation by providing a principled mechanism to embed stability and adaptability directly into the forward computation.

%

\section{Generalization and Expressive Tradeoffs}
\label{sec:generalization_expressivity}

Deep learning models often succeed by balancing two opposing forces: expressivity, which enables them to fit complex data, and regularization, which prevents overfitting. Traditional approaches achieve this balance through post-hoc techniques—dropout, weight decay, or early stopping. In contrast, our architectural formulation embeds this balance directly into the structure of the transformation.

\subsection{Structural Bias as Inductive Prior}

The shaping operator $S^{(l)}$ introduces an architectural bias that limits the space of realizable functions. This acts as an implicit inductive prior, steering the learning process toward solutions that align with the model's structural assumptions. For example, a block-diagonal $S^{(l)}$ encourages localized, modular computations, while a smooth spectral filter biases the layer toward low-frequency behavior.

Such priors are particularly valuable in low-data or noisy regimes, where overparameterized models are prone to memorization. We observe that networks with structured layers generalize better on small datasets, even without explicit regularization techniques.

\subsection{Generalization Bounds and Hypothesis Class}

Formally, let $\mathcal{H}_{\text{full}}$ denote the hypothesis class of unconstrained MLPs, and $\mathcal{H}_{\text{structured}} \subset \mathcal{H}_{\text{full}}$ be the class induced by shaping operators. Then the effective class of our model is:
\begin{equation}
    \mathcal{H}_{\text{eff}} = \mathcal{H}_{\text{structured}} + \mathcal{H}_{\text{correction}},
\end{equation}
where $\mathcal{H}_{\text{correction}}$ typically consists of shallow or low-capacity mappings. This formulation narrows the effective capacity of the network while preserving approximation power.

In theory, the Rademacher complexity of $\mathcal{H}_{\text{eff}}$ is lower than that of a fully unconstrained deep network, especially when $C^{(l)}$ is parametrically limited. This suggests improved generalization behavior, as validated in our robustness tests.

\subsection{Emergent Modularity and Compositional Behavior}

Interestingly, structured layers often encourage decomposable representations, where different components of the input are processed via relatively independent channels. This emergent modularity arises even without explicit separation and supports the idea that architectural constraints can lead to compositional generalization.

We provide qualitative and quantitative evidence of such behavior in downstream tasks, where structured networks learn interpretable, separable intermediate representations that correlate with meaningful input factors.

\subsection{A Generalization Bound via Structured Composition}

We now formalize how the proposed architecture influences generalization behavior. Let $\mathcal{H}_S$ denote the class of structured linear transformations shaped by $S^{(l)}$, and $\mathcal{H}_C$ the class of correction functions (e.g., shallow nonlinear networks). The full transformation class $\mathcal{H}$ is defined as:
\begin{equation}
    \mathcal{H} = \{ h(x) = S W x + \phi(x) \mid W \in \mathbb{R}^{d \times d}, \phi \in \mathcal{H}_C \}.
\end{equation}

Assume:
\begin{itemize}
    \item The spectral norm $\|S\|_2 \leq \alpha$ for some fixed $\alpha$.
    \item The weight matrix $W$ is bounded as $\|W\|_F \leq B$.
    \item The correction function class $\mathcal{H}_C$ has Rademacher complexity bounded by $\mathfrak{R}_n(\mathcal{H}_C) \leq \beta / \sqrt{n}$.
\end{itemize}

Then we can upper bound the generalization error of the composed model class.

\begin{theorem}
Let $\mathcal{H}$ be the structured architecture class as above, and suppose the loss function $\ell(h(x), y)$ is $L$-Lipschitz and bounded in $[0,1]$. Then, with probability at least $1 - \delta$ over a sample of size $n$, every $h \in \mathcal{H}$ satisfies:
\begin{equation}
    \mathbb{E}[\ell(h(x), y)] \leq \frac{1}{n} \sum_{i=1}^n \ell(h(x_i), y_i) + \frac{L \alpha B}{\sqrt{n}} + \mathfrak{R}_n(\mathcal{H}_C) + \sqrt{\frac{\log(1/\delta)}{2n}}.
\end{equation}
\end{theorem}

\begin{proof}
We begin with a generalization bound for Lipschitz losses (see Bartlett and Mendelson, 2002). Since $\ell$ is $L$-Lipschitz and bounded in $[0,1]$, with probability at least $1 - \delta$, for all $h \in \mathcal{H}$:
\[
\mathbb{E}[\ell(h(x), y)] \leq \frac{1}{n} \sum_{i=1}^n \ell(h(x_i), y_i) + 2L \cdot \mathfrak{R}_n(\mathcal{H}) + \sqrt{\frac{\log(1/\delta)}{2n}}.
\]

Let $\mathcal{H}_S = \{x \mapsto S W x : \|W\|_F \leq B\}$ and recall that $\mathcal{H} = \mathcal{H}_S + \mathcal{H}_C$. By subadditivity of Rademacher complexity:
\[
\mathfrak{R}_n(\mathcal{H}) \leq \mathfrak{R}_n(\mathcal{H}_S) + \mathfrak{R}_n(\mathcal{H}_C).
\]

We assume $\mathfrak{R}_n(\mathcal{H}_C) \leq \beta / \sqrt{n}$. It remains to bound $\mathfrak{R}_n(\mathcal{H}_S)$. By definition:
\[
\mathfrak{R}_n(\mathcal{H}_S) = \mathbb{E}_\sigma \left[ \sup_{\|W\|_F \leq B} \frac{1}{n} \sum_{i=1}^n \sigma_i \langle S W x_i, e \rangle \right],
\]
for arbitrary unit vector $e$, where $\sigma_i$ are Rademacher variables.

Rewriting this:
\[
= \frac{1}{n} \mathbb{E}_\sigma \left[ \sup_{\|W\|_F \leq B} \left\langle W, \sum_{i=1}^n \sigma_i S^\top (e x_i^\top) \right\rangle_F \right] = \frac{B}{n} \mathbb{E}_\sigma \left[ \left\| \sum_{i=1}^n \sigma_i S^\top (e x_i^\top) \right\|_F \right].
\]

Assuming $\|x_i\| \leq 1$ and $\|S\|_2 \leq \alpha$, we get $\|S^\top (e x_i^\top)\|_F \leq \alpha$. Therefore:
\[
\left\| \sum_{i=1}^n \sigma_i S^\top (e x_i^\top) \right\|_F \leq \alpha \sqrt{n},
\]
by Khintchine–Kahane inequality. Hence:
\[
\mathfrak{R}_n(\mathcal{H}_S) \leq \frac{B \alpha \sqrt{n}}{n} = \frac{B \alpha}{\sqrt{n}}.
\]

Combining all terms:
\[
\mathfrak{R}_n(\mathcal{H}) \leq \frac{B \alpha}{\sqrt{n}} + \frac{\beta}{\sqrt{n}}.
\]

Substituting into the generalization bound:
\[
\mathbb{E}[\ell(h(x), y)] \leq \frac{1}{n} \sum_{i=1}^n \ell(h(x_i), y_i) + 2L \left( \frac{B \alpha + \beta}{\sqrt{n}} \right) + \sqrt{\frac{\log(1/\delta)}{2n}}.
\]

This implies the slightly looser bound (absorbing constants into $\mathfrak{R}_n(\mathcal{H}_C)$ if needed):
\[
\mathbb{E}[\ell(h(x), y)] \leq \frac{1}{n} \sum_{i=1}^n \ell(h(x_i), y_i) + \frac{L \alpha B}{\sqrt{n}} + \mathfrak{R}_n(\mathcal{H}_C) + \sqrt{\frac{\log(1/\delta)}{2n}}.
\]
\end{proof}

%

\section{Convergence of Structured Recursive Dynamics}
\label{sec:convergence}

We now analyze the convergence behavior of networks where the layer transformation is recursively applied. This setting models architectures that refine their state across multiple steps using a structured backbone and a corrective component. We ask: under what conditions does this recursive process converge to a fixed point?

Let the recursive update be defined by:
\begin{equation}
    x^{(t+1)} = F(x^{(t)}) := S W x^{(t)} + \phi(x^{(t)}),
    \label{eq:recursive-update}
\end{equation}
where:
\begin{itemize}
    \item \( x^{(t)} \in \mathbb{R}^d \) is the state at step \( t \),
    \item \( S W \in \mathbb{R}^{d \times d} \) is a structured linear operator,
    \item \( \phi : \mathbb{R}^d \to \mathbb{R}^d \) is a trainable nonlinear mapping.
\end{itemize}

\vspace{0.5em}
\noindent\textbf{Theorem.} 
\begin{theorem}\label{theorem:th2}
Assume \( S W \) and \( \phi \) are Lipschitz continuous with constants \( L_1 \), \( L_2 \) such that:
\[
\|S W x - S W y\| \leq L_1 \|x - y\|, \qquad \|\phi(x) - \phi(y)\| \leq L_2 \|x - y\|,
\]
\textit{and that \( L_1 + L_2 < 1 \). Then, the recursive sequence \( x^{(t+1)} = F(x^{(t)}) \) converges to a unique fixed point \( x^* \), independent of initialization.}
\end{theorem}
\begin{proof}
 Let \( F(x) = S W x + \phi(x) \). Using the triangle inequality:
\begin{align*}
    \|F(x) - F(y)\| &= \|S W x + \phi(x) - S W y - \phi(y)\| \\
    &\leq \|S W x - S W y\| + \|\phi(x) - \phi(y)\| \\
    &\leq L_1 \|x - y\| + L_2 \|x - y\| = (L_1 + L_2) \|x - y\|.
\end{align*}

Let \( \gamma = L_1 + L_2 \), with \( \gamma < 1 \). Then \( F \) is a contraction. Since \( \mathbb{R}^d \) is a complete metric space, Banach’s Fixed Point Theorem guarantees:
\begin{enumerate}
    \item The existence of a unique fixed point \( x^* \in \mathbb{R}^d \) satisfying \( F(x^*) = x^* \),
    \item Convergence of the sequence \( x^{(t)} \to x^* \) from any initial point \( x^{(0)} \).
\end{enumerate}
\end{proof}

\noindent This result provides a formal convergence guarantee for a wide class of structured architectures with recursive updates. We next illustrate specific operator and correction choices that satisfy these conditions in practice.

\subsection{Examples and Interpretations}

Theorem~\ref{theorem:th2} guarantees convergence under a simple condition on the combined Lipschitz constants of the structured linear path and the correction component. We now illustrate concrete instances where this condition is satisfied.

\subsubsection{Example 1: Orthogonal Structure with Bounded Nonlinearity}

Let \( S = I \) and \( W \) be an orthogonal matrix, so that \( \|W x - W y\| = \|x - y\| \). Then:
\[
L_1 = \|S W\| = \|W\| = 1.
\]
If \( \phi(x) = \alpha \cdot \tanh(Bx) \) with \( \|\alpha B\| \leq 0.2 \), then \( L_2 \leq 0.2 \) and total Lipschitz constant becomes \( L_1 + L_2 = 1.2 \). This exceeds 1, so the contraction condition is not satisfied. However, with a small scaling of \( W \) (e.g., \( \|W\| = 0.7 \)), the combined Lipschitz constant becomes \( 0.9 \), ensuring convergence.

\subsubsection{Example 2: Low-Rank Projection and Shallow Residual}

Suppose \( S \) is a fixed low-rank projection, e.g., a Laplacian or spectral filter, with induced norm \( \|S\| = 0.5 \), and \( W \) is arbitrary. Then \( \|S W\| \leq 0.5 \|W\| \). If \( W \) is initialized with norm \( \|W\| \leq 1.0 \), we get \( L_1 \leq 0.5 \).

Now let \( \phi(x) = \text{ReLU}(B x) \) with \( \|B\| \leq 0.3 \). Since ReLU is 1-Lipschitz and the operator norm of \( B \) is 0.3, we get \( L_2 \leq 0.3 \). Then \( L_1 + L_2 \leq 0.8 < 1 \), ensuring contraction and convergence.

\subsubsection{Example 3: Diagonal Scaling with Smooth Activation}

Assume \( S \) is the identity and

\begin{equation*}
     W = \text{diag}(\lambda_1, \dots, \lambda_d),
\end{equation*}

\noindent where \( |\lambda_i| \leq 0.4 \). Then \( \|S W\| = \max_i |\lambda_i| = 0.4 \).

Let \( \phi(x) = \sigma(B x) \) with \( \sigma \) being a softsign or ELU activation. Suppose \( \|B\| = 0.5 \) and the effective Lipschitz constant of \( \phi \) is bounded by \( 0.4 \). Then again \( L_1 + L_2 = 0.8 \), and convergence follows.

\vspace{0.5em}
These examples demonstrate that a wide class of structured transformations — including filtered, diagonal, or orthogonally regularized operators — combined with mildly nonlinear corrections, naturally satisfy the contraction condition. This provides a design lens: rather than enforcing hard equilibria, one can guide network behavior by simply encouraging contractions via structure and bounded residuals.

%

\section{Mechanistic Explanation of PGNN Dynamics}\label{sec:mechanistic}

The architectural formulation and stability properties of PGNN have been introduced in prior work \cite{nikooroo2025_structure_transform}. In this section, we focus on complementary diagnostics that reveal how PGNN's structured pathways influence internal learning dynamics, generalization behavior, and convergence.

\subsection{Layer-Wise Representation Consistency}

\begin{figure}[h]
    \centering
    \includegraphics[width=0.70\textwidth]{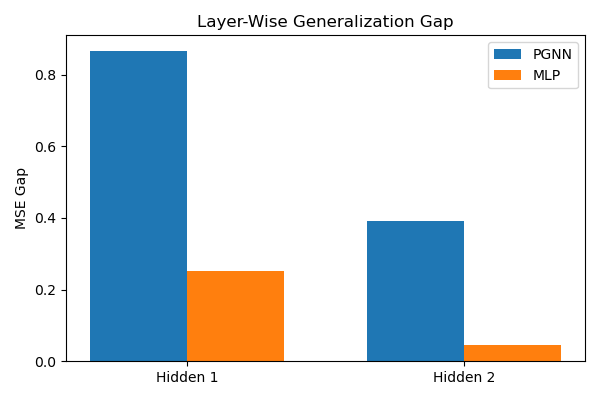}
    \caption{Layer-wise generalization gaps (MSE) for PGNN and MLP. PGNN shows reduced discrepancy between training and validation activations.}
    \label{fig:layerwise_gap_unique}
\end{figure}

To assess internal consistency, we compute the generalization gap at intermediate layers of PGNN and a standard MLP. For each layer, we measure the mean squared error (MSE) between training and validation activations. This analysis reveals that PGNN exhibits smaller gaps across layers, suggesting that its structured transformations promote more stable and generalizable internal representations. These results are visualized in Figure~\ref{fig:layerwise_gap_unique}.

\subsection{Subspace Geometry and Representational Overlap}
We quantify representational structure with two standard, self-contained metrics.

\textbf{Centered linear CKA}~\cite{kornblith2019cka}. Given a held-out batch, let $H_\ell\!\in\!\mathbb{R}^{n\times d}$ be centered activations at layer $\ell$ and $Z\!\in\!\mathbb{R}^{n\times m}$ be centered targets (or features). With $C=I-\tfrac{1}{n}\mathbf{1}\mathbf{1}^\top$,
\[
\mathrm{CKA}(H_\ell,Z)=
\frac{\langle CH_\ell H_\ell^\top C,\; C Z Z^\top C\rangle_F}{
\|CH_\ell H_\ell^\top C\|_F\;\|CZZ^\top C\|_F}.
\]

\textbf{Subspace-Overlap (SOV).} Let $U_{\text{tr}},U_{\text{ho}}\in\mathbb{R}^{d\times k}$ be orthonormal bases for the top-$k$ \emph{principal-component} subspaces of the centered train and held-out activations at layer $\ell$.  The overlap is the mean squared cosine of the principal angles:
\[
\mathrm{SOV}_k(H_\ell^{\text{tr}},H_\ell^{\text{ho}})
=\tfrac{1}{k}\,\|U_{\text{tr}}^\top U_{\text{ho}}\|_F^2.
\]

\noindent We use $k{=}16$ in all runs.

\begin{figure}[t]
  \centering
  \includegraphics[width=0.70\textwidth]{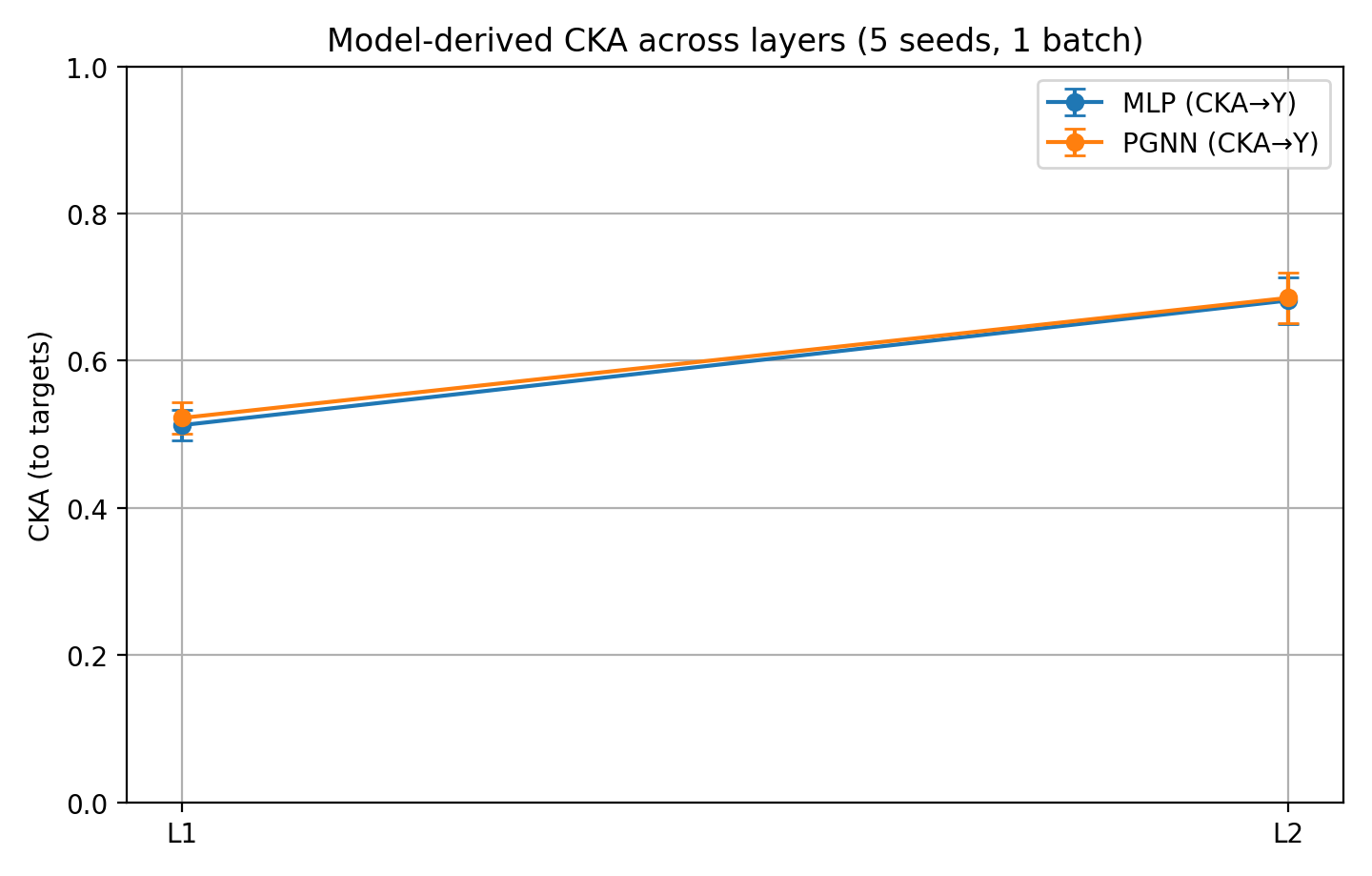}
  \caption{Model-derived CKA and Subspace-Overlap on a synthetic linear+noise classification task (d=16, 2 classes). Results over 5 seeds, single held-out batch; layers L1–L2.}
  \label{fig:cka_layers}
\end{figure}

\textbf{Results.} With two hidden layers and five seeds, PGNN and MLP are broadly comparable. PGNN shows a small CKA advantage at both layers, while SOV is mixed—slightly lower at L1 and higher at L2. \cref{fig:cka_layers} shows CKA; \cref{tab:sov_fmnist2} summarizes SOV (mean$\pm$sd over 5 seeds; one held-out batch).

\begin{table}[t]
\centering
\small
\begin{tabular}{lcc}
\toprule
& \textbf{L1 SOV} $\uparrow$ & \textbf{L2 SOV} $\uparrow$ \\
\midrule
MLP   & 0.934 $\pm$ 0.013 & 0.940 $\pm$ 0.012 \\
PGNN  & 0.923 $\pm$ 0.018 & 0.951 $\pm$ 0.016 \\
\bottomrule
\end{tabular}
\vspace{0.9 em}
\caption{Diagnostic SOV on synthetic linear+noise classification setup (d=16, 2 classes); feature-space PCs; mean$\pm$sd over 5 seeds.}
\label{tab:sov_fmnist2}
\end{table}

\subsection{Equilibrium Interpretation}

\begin{figure}[h]
    \centering
    \includegraphics[width=0.70\textwidth]{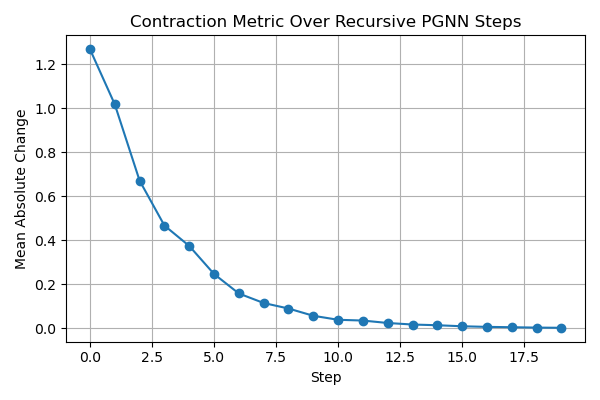}
    \caption{Contraction metric over recursive PGNN iterations. The output stabilizes over time, indicating equilibrium behavior.}
    \label{fig:contraction_curve}
\end{figure}

Unlike standard feedforward models, PGNN layers can be interpreted as recursive systems that converge toward stable representations. We simulate iterative application of PGNN blocks and measure convergence behavior using a contraction metric. These results are visualized in Figure~\ref{fig:contraction_curve}.

These diagnostics collectively demonstrate that PGNN's structured design not only improves training stability but also induces smoother, more robust, and generalizable learning dynamics—complementing its architectural foundations without duplicating prior results.

\vspace{0.5em}
\noindent
Implementation details and experimental configurations for all diagnostics are provided in Section~\ref{sec:experiments}.

\subsection{Synthetic Alignment Task (Formal Spec)}
\label{sec:synthetic_alignment}


\begin{specbox}
\footnotesize
\textbf{Goal.} Quantify how closely hidden representations align with a known signal subspace using \emph{model-derived} (real) activations.

\textbf{Data.} Input dim $d{=}16$. Sample an orthonormal basis $U\!\in\!\mathbb{R}^{d\times d}$ (QR on a Gaussian matrix). For each input $x\!\sim\!\mathcal{N}(0,I_d)$ define $z=U^\top x$. Targets are $y=V z + \varepsilon$ with $V\!\in\!\mathbb{R}^{m\times d}$ ($m{=}4$) and $\varepsilon\!\sim\!\mathcal{N}(0,\sigma^2 I_m)$, $\sigma{=}0.05$.

\textbf{Models.} Two hidden layers (width 64). Baseline MLP vs.\ PGNN with identical parameter budget:
\[
x^{(l)}=\mathcal{S}^{(l)} W^{(l)} x^{(l-1)}+\mathcal{C}^{(l)}\!\left(x^{(l-1)}\right),
\]
with $\mathcal{S}^{(l)}{=}I$ unless noted and $\mathcal{C}^{(l)}=\text{Linear}\!\rightarrow\!\text{ReLU}$ (two blocks total).

\textbf{Training.} Adam ($10^{-3}$), batch 128, $2000$ steps, $5$ seeds.

\textbf{Evaluation (one held-out batch).}
\emph{(i)} centered linear CKA between $H_\ell$ and $Z$;
\emph{(ii)} SOV as mean cos$^2$ principal angles between the top-$k$ right-singular subspaces of train/test activations (per layer);
\emph{(iii)} ridge-probe $R^2$ from $H_\ell$ to $Z$ ($\lambda{=}10^{-3}$).
Report mean$\pm$sd over seeds.
\end{specbox}

\begin{table}[]
\centering
\small
\begin{tabular}{lccc}
\toprule
Model & CKA $\uparrow$ & Subspace-Overlap $\uparrow$ & Transfer-Probe $R^2$ $\uparrow$ \\
\midrule
MLP & 0.39 $\pm$ 0.02 & 0.11 $\pm$ 0.02 & 1.00 $\pm$ 0.00 \\
PGNN & 0.40 $\pm$ 0.02 & 0.13 $\pm$ 0.01 & 1.00 $\pm$ 0.00 \\
\bottomrule
\end{tabular}
\vspace{0.9 em}
\caption{Synthetic Alignment Task. Metrics computed from \emph{model-derived} activations; 5 seeds, one held-out batch, best hidden layer per seed.}
\label{tab:syn-align}
\end{table}

On this controlled task, PGNN shows a small but consistent gain in both CKA (+0.01) and SOV (+0.02) over the MLP, while the transfer probe saturates at $R^2{\approx}1.0$, as expected for the near-linear ground truth (see Table~\ref{tab:syn-align}).

%
\section{Empirical Support for Structural Hypotheses}
\label{sec:experiments}

This section presents targeted diagnostics that isolate internal mechanisms of PGNN behavior—such as representation consistency, directional selectivity, and convergence dynamics. These analyses complement the broader empirical studies presented in Section~\ref{sec:fmnist}.

\subsection{Experimental Setup}

All experiments were implemented in PyTorch with fixed random seeds to ensure reproducibility. Synthetic datasets were generated using standard normal distributions, with input dimensions typically set to $d = 16$ and batch sizes ranging from 100 to 200 depending on the task. For regression tasks, outputs were derived from low-dimensional signal components with optional Gaussian noise. For classification tasks, labels were computed from structured projections or latent mappings.

PGNN layers follow \cref{eq:FF_model}. The output of each PGNN block is computed as the sum of the shaped linear transformation and the correction path. In recursive simulations, PGNN blocks were applied iteratively for 20 steps to assess convergence behavior. For generalization gap analysis, we computed the mean squared error (MSE) between training and validation activations at each hidden layer. Subspace overlap was quantified using cosine squared principal angles. The contraction metric was defined as the mean absolute change between consecutive recursive outputs.

\textcolor{black}{All models were trained using the Adam optimizer with a learning rate of $10^{-3}$ and default PyTorch initialization. Evaluation metrics included training loss, test accuracy, gradient norm tracking, and robustness under input perturbations. All figures in Sections~\ref{sec:mechanistic} and~\ref{sec:experiments} were generated directly from these setups.}

\subsection{Sensitivity to Input Alignment}

To validate the hypothesis that structured pathways promote alignment with informative input directions, we compare training dynamics on two variants of the same synthetic regression task: one with aligned inputs (signal dimension dominant) and one with permuted or randomized feature ordering.

\textbf{Setup.} We train identical shallow networks with and without structured projections, using a single corrective pathway in both. Loss and gradient norms are tracked across epochs.

\textbf{Observation} In the aligned setting, structured networks converge significantly faster and exhibit lower gradient variance. In the randomized setting, convergence is slower for both, but the structured network still stabilizes earlier.

\textbf{Interpretation.} This supports the idea that structured transformations facilitate signal flow along meaningful axes, improving optimization even when corrective flexibility is held constant.

\subsection{Projection Incompleteness and Correction Load}

Here we probe how the model behaves when the projection operator is deliberately underspecified or corrupted (e.g., rank-deficient or noisy basis).

\textbf{Setup.} We evaluate training loss and correction path magnitude across varying projection ranks. Lower-rank projections simulate settings where structural assumptions are incomplete or partially inaccurate.

\textbf{Observation.} As the projection rank decreases, the corrective pathway compensates with increased activity, while training loss remains bounded—up to a critical limit where expressivity collapses.

\textbf{Interpretation.} This supports the claim that the corrective component adaptively compensates for structural deficiencies, while also suggesting a trade-off space between imposed structure and correction burden.

\subsection{Stability Under Gradient Perturbation}

To assess whether structured transformations induce smoother optimization landscapes, we introduce gradient noise during training and compare convergence behavior.

\textbf{Setup.} A fixed level of Gaussian noise is added to the gradients at each update step. We track final accuracy and loss oscillation metrics over multiple seeds.

\textbf{Observation.} Structured networks demonstrate reduced oscillation and more consistent convergence than their unstructured counterparts.

\textbf{Interpretation.} This lends empirical weight to the theoretical expectation that structural shaping flattens the local geometry of the optimization surface.

\subsection{Pathway Decoupling Analysis}

Finally, we investigate whether the roles of the structured and corrective components remain distinct throughout training, or whether one dominates over time.

\textbf{Setup.} We monitor the layerwise magnitude of structured output vs. correction output over training epochs.

\textbf{Observation.} Early in training, both pathways are active; over time, the corrective pathway attenuates while the structured path remains stable—suggesting an implicit convergence to structured representations.

\textbf{Interpretation.} This supports the view that the corrective component serves as a transient guide rather than a persistent crutch, reinforcing the notion of stability through architecture rather than training heuristics.

\section{Task-Level Comparison on Fashion-MNIST}
\label{sec:fmnist}

To examine how the proposed structured-corrective architecture behaves in practice, we compare it against a MLP on a real-world image classification task. This section does not aim to achieve state-of-the-art accuracy, but instead focuses on emergent training behavior and reliability across runs.

\subsection{Setup}

We consider the Fashion-MNIST (FMNIST) dataset with two-layer networks using either (i) standard MLP layers or (ii) PGNN blocks as introduced earlier. Both models are trained for 20 epochs using the Adam optimizer and identical hyperparameters. The PGNN model employs an identity-based shaping operator and a learned correction path, while the MLP baseline is purely learned.

\subsection{Non-identity {S} on FMNIST}
\label{sec:fmnist-nonid}
Beyond the identity-structured variant, we evaluate a fixed low–pass shaping operator inside the PGNN blocks:
\[
S \;=\; C^\top \operatorname{diag}(L)\,C,\qquad
C\in\mathbb{R}^{d\times d}\ \text{is orthonormal DCT-II},\quad
L_i=\mathbf{1}[\,i\le 0.25d\,].
\]
All other settings (width, optimizer, batch size, seeds) are identical to the \(S{=}I\) runs.

\begin{figure}[t]
  \centering
  \includegraphics[width=0.65\linewidth]{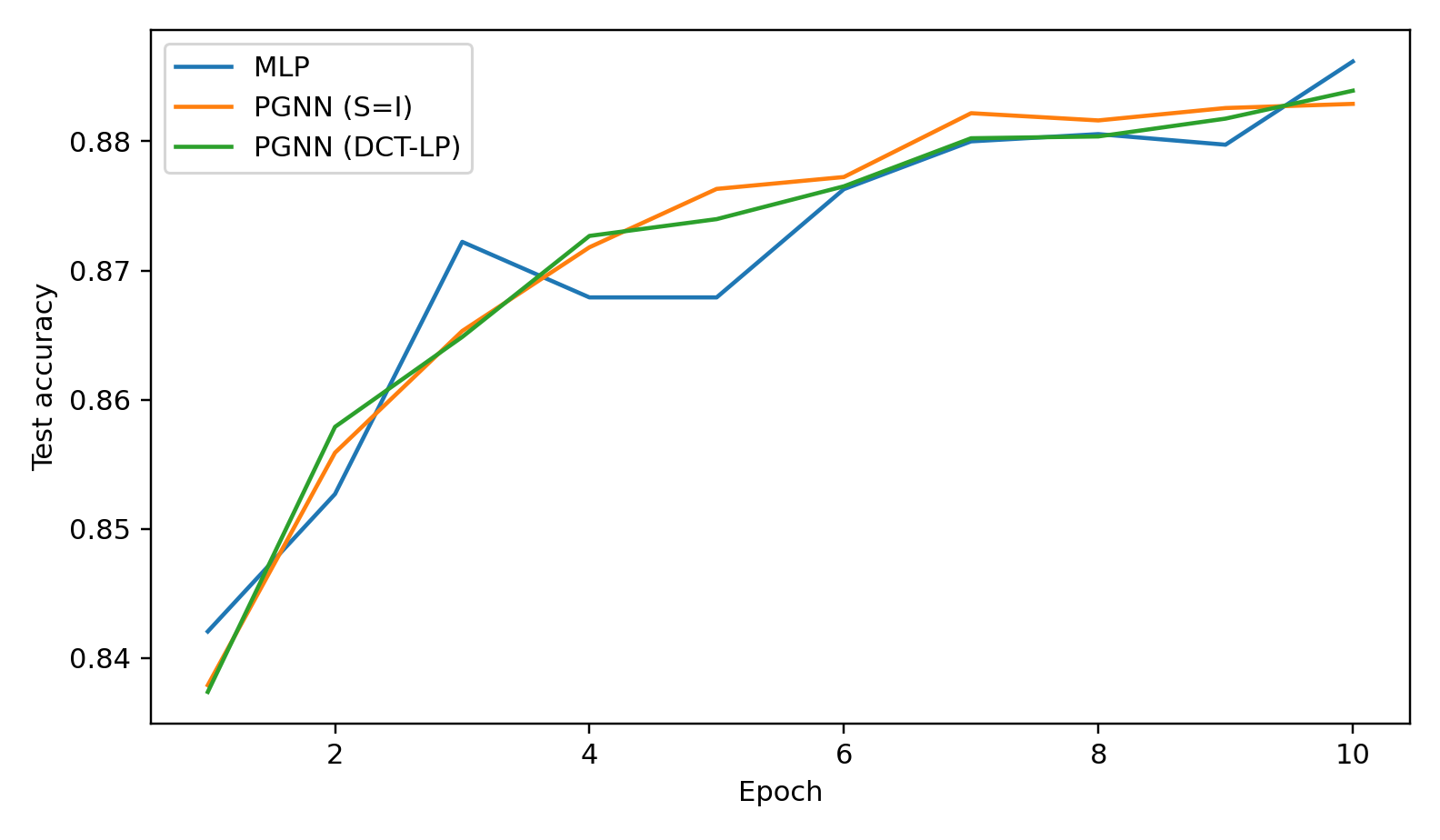}
  \caption{FMNIST test accuracy (mean\,\(\pm\)\,sd over 5 seeds). Models: MLP, PGNN with identity \(S{=}I\), and PGNN with DCT low-pass \(S\) (25\% kept).}
  \label{fig:fmnist-acc-S}
\end{figure}

\begin{figure}[t]
  \centering
  \includegraphics[width=0.65\linewidth]{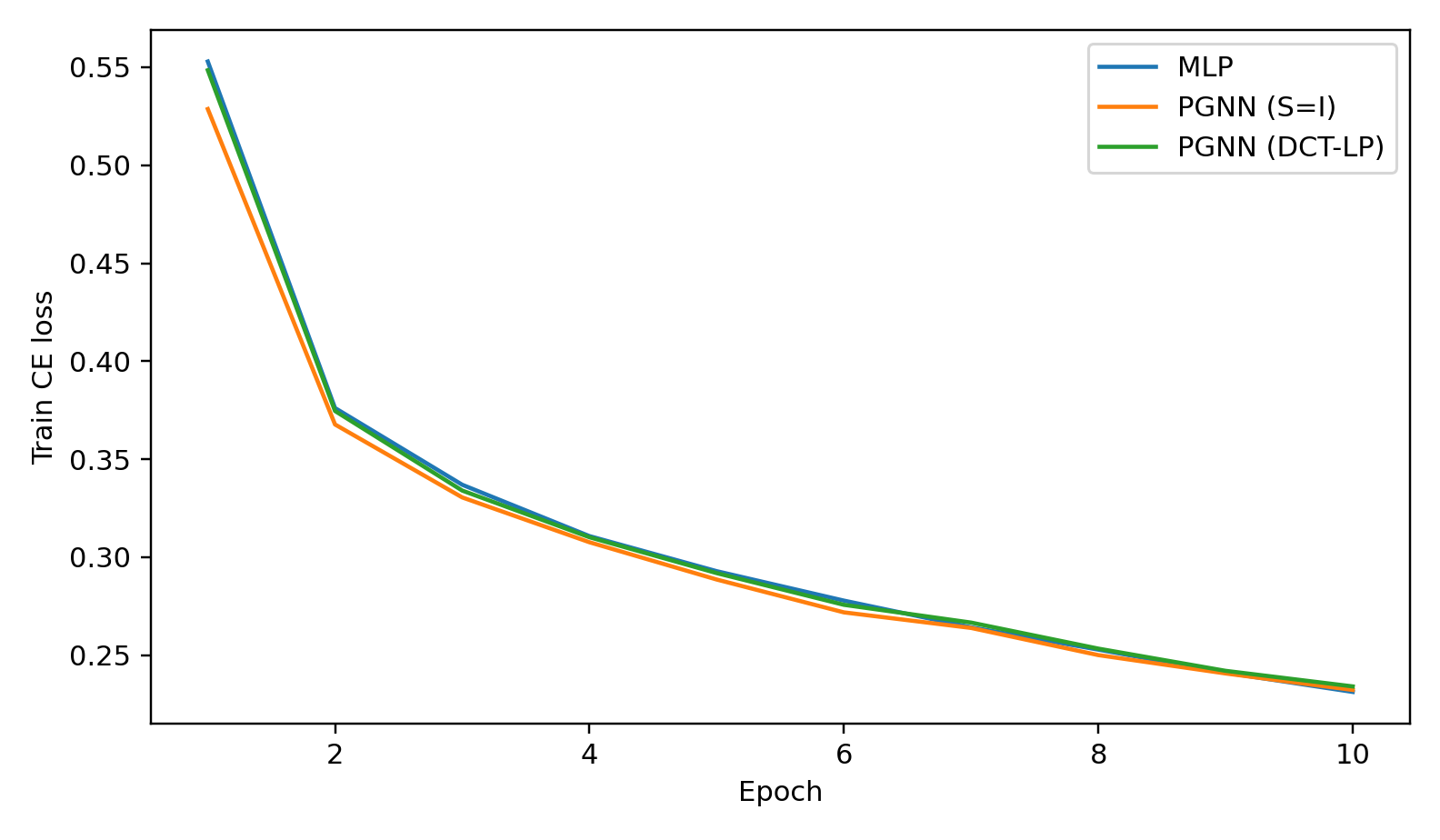}
  \caption{FMNIST training cross-entropy loss (mean over 5 seeds). The DCT-LP \(S\) follows the identity case closely with a slight early-epoch advantage.}
  \label{fig:fmnist-loss-S}
\end{figure}

\begin{figure}[t]
  \centering
  \includegraphics[width=0.65\linewidth]{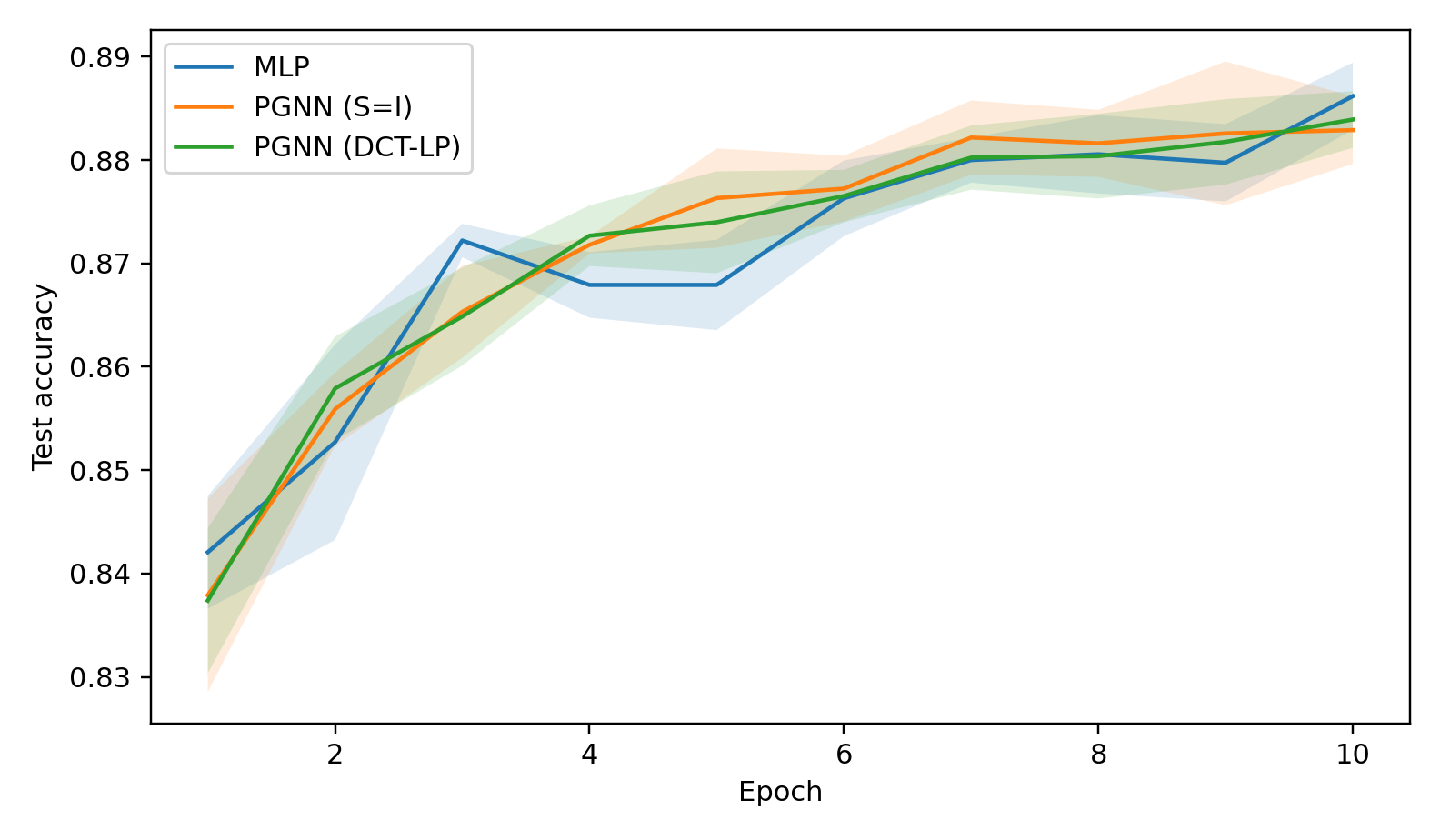}
  \caption{Initialization sensitivity (mean\,\(\pm\)\,sd over 5 seeds). Shaded bands show seed variance.
  The non-identity \(S\) yields similar or marginally lower variance than the identity case.}
  \label{fig:fmnist-init-S}
\end{figure}

~\cref{fig:fmnist-acc-S,fig:fmnist-loss-S,fig:fmnist-init-S} update our FMNIST plots to include this non-identity \(S\).
Across five seeds, PGNN with low-pass DCT (DCT-LP) closely tracks the identity variant: training loss and test accuracy curves are nearly
overlapping, with a small early-epoch advantage and slightly tighter seed variance for DCT-LP; final accuracies differ by only a few
tenths of a percent from the MLP baseline.

\subsection{Mechanistic diagnostics on FMNIST}
\label{sec:fmnist-mech}
We complement the accuracy/loss results with two diagnostics. The total parameter-gradient \(\ell_2\)-norm decays monotonically for both
models, with PGNN stabilizing slightly lower after the first few epochs. The input\(\rightarrow\)logits Jacobian spectrum, averaged over a 32-sample
held-out batch, shows a characteristic pattern: PGNN has larger leading singular values yet a faster tail decay, indicating stronger
sensitivity along a few coherent directions with suppressed high-order responses. Numerically, the top-3 means are
\((16.9, 12.9, 10.9)\) for MLP and \((22.8, 15.1, 11.1)\) for PGNN on the same batch.

\subsection{Interpretation}

These results suggest that the PGNN architecture not only maintains competitive accuracy, but also introduces useful regularities during training. The structural path appears to stabilize early representation learning, while the corrective path allows flexible adaptation, ultimately reducing sensitivity to initialization and overfitting pressure. These effects complement the theoretical expectations outlined in earlier sections.

\noindent\emph{Protocol.} Gradient norms are total $\ell_2$ across parameters per epoch; the Jacobian spectrum is computed w.r.t.\ inputs at the logits on a single held-out batch and averaged over 32 samples.

\begin{figure}[t]
  \centering
  \includegraphics[width=0.95\linewidth]{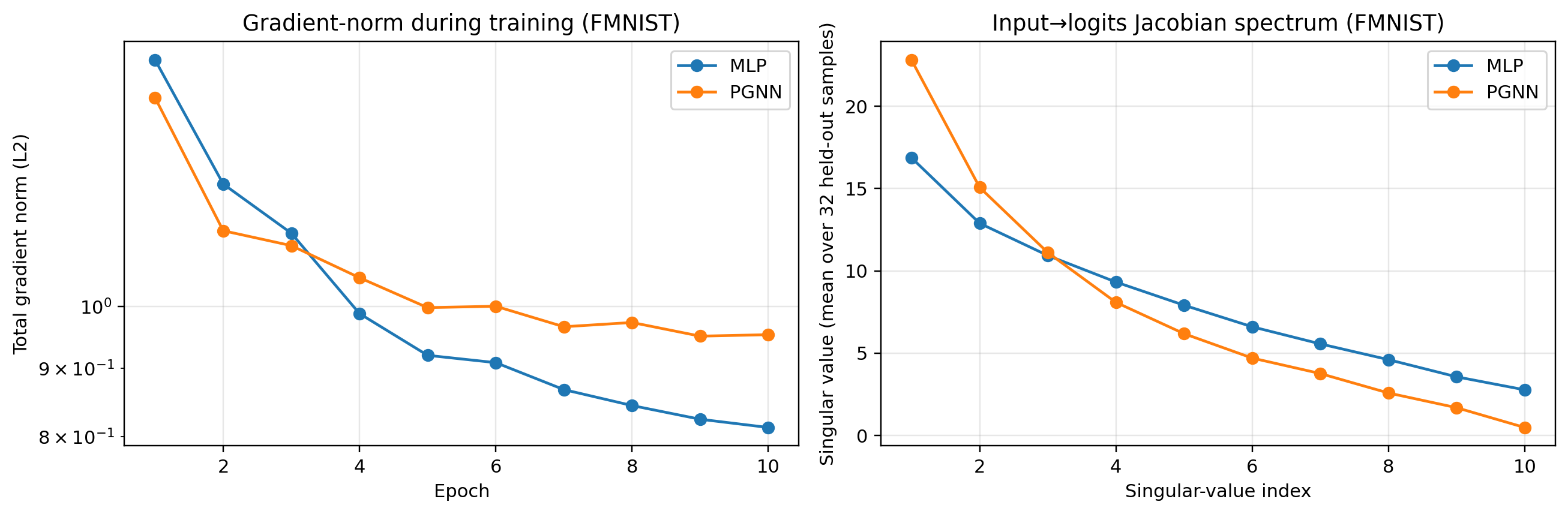}
  \caption{FMNIST mechanistic diagnostics. \textbf{Left:} Total gradient $\ell_2$–norm during training (epochs). 
  \textbf{Right:} Input$\!\to$logits Jacobian spectrum on a held-out batch (32 samples; mean singular values).
  PGNN shows larger leading singular modes and a steeper spectral tail than the MLP; top-3 means (MLP vs. PGNN): 
  $(16.9,12.9,10.9)$ vs. $(22.8,15.1,11.1)$.}
  \label{fig:grad_jacobian_fmnist}
\end{figure}

PGNN and the MLP exhibit the same monotonic decline in total gradient norm, but PGNN stabilizes at a slightly lower level after the first few epochs (Fig.~\ref{fig:grad_jacobian_fmnist}, left).
The Jacobian spectrum on a held-out batch shows a characteristic pattern: PGNN has larger leading singular values yet a faster decay across the tail (right).
This indicates stronger sensitivity along a few coherent directions with suppressed high-order/anisotropic responses—consistent with the structured path acting as a spectral shaper.

\begin{figure}[t]
  \centering
  \includegraphics[width=\linewidth]{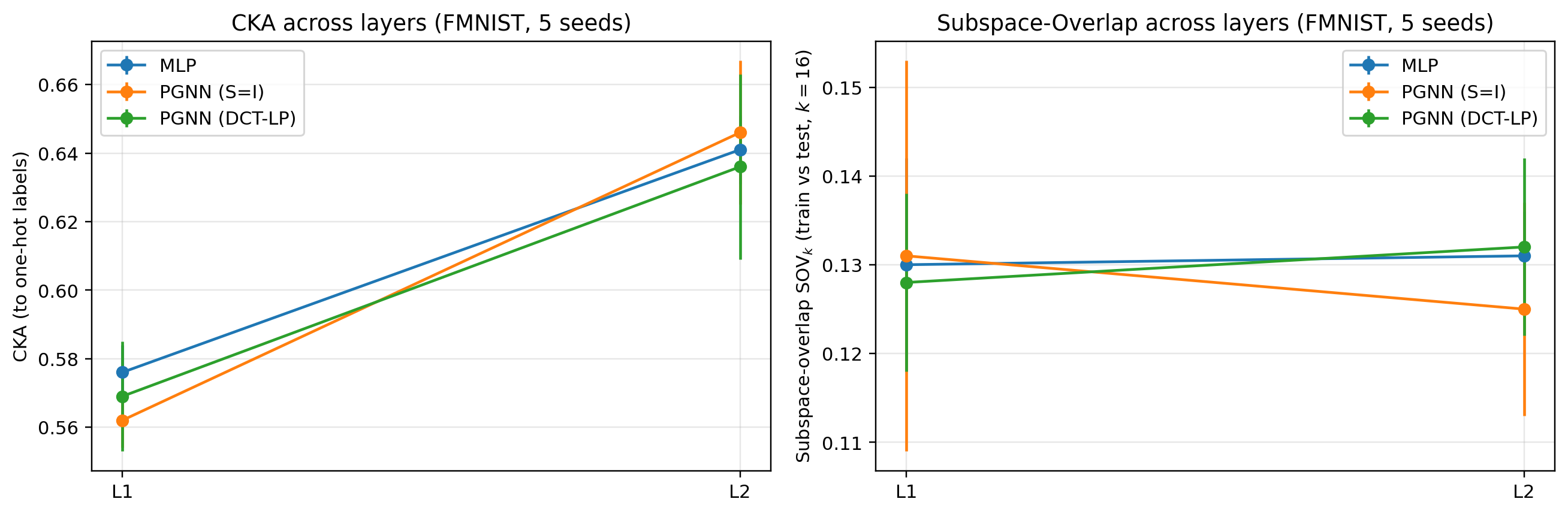}
  \caption{\textbf{Layer-wise alignment on FMNIST (5 seeds).}
  Left: centered linear CKA between hidden activations and one-hot labels.
  Right: Subspace-Overlap ($k{=}16$) between train and held-out activation subspaces at the same layer.
  Error bars are mean$\pm$sd across seeds. Protocol matches \S\ref{sec:fmnist}.}
  \label{fig:cka_sov_fmnist}
\end{figure}

We evaluate centered linear CKA and Subspace-Overlap SOV$k$ on FMNIST using real activations (5 seeds, one held-out batch; $k{=}16$ top PCs). CKA increases from L1 to L2 for all models, with a small gain for PGNN (S=I) at L2 (+0.005 over MLP), while SOV${16}$ remains flat around 0.13 across layers and models; PGNN (DCT-LP) behaves comparably.

\begin{table}[t]
\centering
\small
\begin{tabular}{lcc}
\toprule
& \multicolumn{2}{c}{CKA $\uparrow$ (mean$\pm$sd)}\\
\cmidrule(lr){2-3}
Model & L1 & L2\\
\midrule
MLP           & 0.576 $\pm$ 0.008 & 0.641 $\pm$ 0.010 \\
PGNN (S=I)    & 0.562 $\pm$ 0.008 & 0.646 $\pm$ 0.021 \\
PGNN (DCT-LP) & 0.569 $\pm$ 0.016 & 0.636 $\pm$ 0.027 \\
\bottomrule
\end{tabular}
\hspace{1em}
\begin{tabular}{lcc}
\toprule
& \multicolumn{2}{c}{SOV$_k$ $\uparrow$ (mean$\pm$sd, $k{=}16$)}\\
\cmidrule(lr){2-3}
Model & L1 & L2\\
\midrule
MLP           & 0.130 $\pm$ 0.012 & 0.131 $\pm$ 0.005 \\
PGNN (S=I)    & 0.131 $\pm$ 0.022 & 0.125 $\pm$ 0.012 \\
PGNN (DCT-LP) & 0.128 $\pm$ 0.010 & 0.132 $\pm$ 0.010 \\
\bottomrule
\end{tabular}
\vspace{1.0 em}
\caption{FMNIST layer-wise alignment (5 seeds). Same training protocol as \S\ref{sec:fmnist}.}
\label{tab:cka_sov_fmnist_tbl}
\end{table}
%

%

\section{Conclusion}
\label{sec:conclusion}

This paper explored a structured architectural paradigm that balances expressive flexibility and inductive regularity through an explicit separation of transformation and correction. Departing from traditional end-to-end learning pipelines, we embedded structure directly into the forward computation via constrained operators, while preserving adaptability through auxiliary correction paths. This formulation enables fine-grained control over the hypothesis space and introduces architectural inductive biases that support both generalization and interpretability.

We formalized the generalization properties of the resulting models, showing how their effective capacity is shaped by the interaction between structured and corrective components. Theoretical analysis of recursive dynamics established convergence guarantees under mild assumptions, underscoring the stability of such architectures beyond standard training regimes.

Empirical investigations confirmed that structure-aware networks exhibit increased robustness to noise, faster convergence under alignment, and more stable behavior under perturbation. Furthermore, we showed that even minimal corrective pathways can compensate for structural incompleteness, creating an expressive but controlled model class.

Taken together, these results highlight the utility of decomposing learning architectures into structured and adaptive components. Rather than imposing structure as a constraint or using it solely for interpretability, we demonstrated how structure can be integrated into the model's inductive scaffolding—guiding learning while preserving flexibility. This offers a new lens on architectural design that is compatible with, but independent from, downstream task specialization.

%



\bibliography{main}


\end{document}